\pdfoutput=1

\documentclass[11pt]{article}

\usepackage{acl}

\usepackage{times}
\usepackage{latexsym}

\usepackage[T1]{fontenc}

\usepackage[utf8]{inputenc}

\usepackage{microtype}

\usepackage{latexsym}
\usepackage{graphicx}
\usepackage{caption}
\usepackage{subcaption}
\usepackage{booktabs}
\usepackage{amsmath,amsfonts,amssymb,amsthm}
\usepackage{array}
\usepackage{makecell}
\usepackage{placeins}
\usepackage{tabularx}
\usepackage{adjustbox}
\usepackage{float}
\usepackage{longtable}
\usepackage{multirow}
\usepackage{xurl}

\usepackage{geometry}
\usepackage{bm}
\usepackage{commath}
\usepackage{enumerate}
\usepackage{accents}
\usepackage[shortlabels]{enumitem}
\usepackage{dsfont}
\usepackage{mathtools}
\usepackage{physics}
\usepackage{bbm}
\usepackage{longtable}
\usepackage[linesnumbered,ruled]{algorithm2e}

\usepackage[margins]{trackchanges}

\makeatletter
\newcommand*\bigcdot{\mathpalette\bigcdot@{.5}}
\newcommand*\bigcdot@[2]{\mathbin{\vcenter{\hbox{\scalebox{#2}{$\m@th#1\bullet$}}}}}
\makeatother

\title{A Bayesian approach to modeling topic-metadata relationships}

\author{Patrick Schulze$^1$\textsuperscript{$\clubsuit$} \And 
        Simon Wiegrebe$^1$\textsuperscript{$\spadesuit$}\\ \\
        \hspace{5cm}$^1$ Department of Statistics, Ludwig-Maximilians-Universität, Munich, Germany \\ 
        \hspace{5cm}$^2$ Department of Political Science, Ludwig-Maximilians-Universität, Munich, Germany \\
        \hspace{5cm}\small \textsuperscript{$\clubsuit$}\texttt{pa.schulze@campus.lmu.de}, \quad \textsuperscript{$\spadesuit$}\texttt{\{simon.wiegrebe,chris,matthias\}@stat.uni-muenchen.de},\\ 
        \hspace{5cm}\small \textsuperscript{$\diamondsuit$}\texttt{paul.thurner@gsi.uni-muenchen.de}
        \And Paul W. Thurner$^2$\textsuperscript{$\diamondsuit$} \\ 
        \hspace{-4.5cm} \textbf{Matthias Aßenmacher}$^1$\textsuperscript{$\spadesuit$}
        \And Christian Heumann$^1$\textsuperscript{$\spadesuit$} }

\begin{document}

\addeditor{MA}
\addeditor{SW}


\maketitle
\begin{abstract}
The objective of advanced topic modeling is not only to explore latent topical structures, but also to estimate relationships between the discovered topics and theoretically relevant metadata. Methods used to estimate such relationships must take into account that the topical structure is not directly observed, but instead being estimated itself in an unsupervised fashion, usually by common topic models.
A frequently used procedure to achieve this is the \textit{method of composition}, a Monte Carlo sampling technique performing multiple repeated linear regressions of sampled topic proportions on metadata covariates.
In this paper, we propose two modifications of this approach: First, we substantially refine the existing implementation of the method of composition from the \textsf{R} package \texttt{stm} by replacing linear regression with the more appropriate Beta regression. Second, we provide a fundamental enhancement of the entire estimation framework by substituting the current blending of frequentist and Bayesian methods with a fully Bayesian approach. This allows for a more appropriate quantification of uncertainty. We illustrate our improved methodology by investigating relationships between Twitter posts by German parliamentarians and different metadata covariates related to their electoral districts, using the Structural Topic Model to estimate topic proportions.
\end{abstract}

\section{Introduction}
\label{intro}

The rise of social media has led to an unprecedented increase in the supply of publicly available unstructured text data. Researchers often wish to examine relationships between observable metadata (e.g., characteristics of a document's author) and in-text patterns \citep{farrell2016corporate, kim2017political}. Probabilistic topic models identify such in-text patterns by producing a posterior distribution over different topics. Yet estimating relationships with observed metadata is not trivial as the target variable is latent and itself being estimated from the text data. In this work we focus on exploring and estimating relationships between metadata and topics learned by the Structural Topic Model \cite[STM;][]{roberts2016model}. We selected this model due to its high relevance in the social sciences - see Appendix \ref{a:implausible}.\footnote{However, it is crucial to understand that the choice of the topic model is only relevant for the estimation of topic proportions and does not affect the methodology for subsequent estimation of topic-metadata relationships. Therefore, the contributions presented in this work are equally valid and applicable when other topic models - such as the Latent Dirichlet Allocation \cite[LDA;][]{blei2003latent} or the Correlated Topic Model \cite[CTM;][]{blei2007correlated} - are used for the initial estimation of topic proportions.} The \textsf{R} package \texttt{stm} \citep{stm} implements the STM itself and additionally provides a framework for estimating topic-metadata relationships via the \textit{method of composition}, a combination of Monte Carlo sampling and frequentist linear regression. Even though this estimation technique is prone to producing predictions incompatible with standard definitions of probability, it is frequently applied in the literature (cf. Appendix \ref{a:implausible}). This leads to implausibilities of two different forms: authors sometimes report negative expected topic proportions \citep[e.g. ][see also our Fig. \ref{fig:estimateEffect}]{farrell2016corporate,moschella2019central}; whereas in other cases "only" the confidence bands partly include negative values \citep[e.g.][]{cho2017anchoring,chandelier2018content,bohr2018key,heberling2019changing}. In both cases, it is ignored that sampled topic proportions are confined to $(0,1)$ by definition, which severely harms the interpretability of results.

In this paper, we suggest two key modifications to the \texttt{stm} implementation in \textsf{R} \citep{stm}: First, our proposed Beta regression approach is a natural correction of the linear regression approach, accounting for topic proportions being restricted to the interval $(0,1)$. Second, we develop a \textit{Bayesian} design within the \textit{method of composition} to allow for a more coherent estimation and interpretation of topic-metadata relationships; in particular, we obtain a posterior predictive distribution of topic proportions at different values of metadata covariates.

We demonstrate the added value of our corrections by analyzing Twitter posts of German politicians, gathered from September 2017 through April 2020. Politics has been particularly impacted by the increasing usage of social media as evidenced by the Brexit vote and US presidential elections, with Twitter being extensively used for direct communication by politicians. We investigate relationships between latent topics in the tweets of German members of parliament (MPs) and corresponding metadata, such as tweet date or unemployment rate in the respective MP's electoral district. In doing so, we attempt to link the topics discussed to specific events as well as to socioeconomic characteristics of the MP's electoral districts.

\section{Background}
\label{Background}

Topic models seek to discover latent thematic clusters, called topics, within a collection of discrete data, usually text documents. In addition to identifying such clusters, topic models estimate the proportions of the discovered topics within each document. Many topic models build upon the well-known LDA, which is a generative probabilistic three-level hierarchical Bayesian mixture model that assumes a Dirichlet distribution for topic proportions. The Correlated Topic Model \cite[CTM;][]{blei2007correlated}, for instance, builds on the LDA but replaces the Dirichlet distribution with a logistic normal distribution in order to capture inter-topic correlations. The STM adopts this approach, but additionally incorporates document-level metadata into the estimation of topics:\footnote{Within the STM, document-level covariates can also be used to fine-tune topic-word distributions \citep{roberts2016model}, but we do not further discuss this here.}

\begin{itemize}
    \item For each document, indexed by $d \in \{1,\dots,D\}$, and each topic, indexed by $k \in \{1,\dots,K\}$, a topic proportion $\theta_{d,k}$ is drawn from a logistic normal distribution.\footnote{The \texttt{stm} package provides several metrics to choose the hyperparameter \textit{K}, as will be discussed in Section \ref{Model Fitting and Global-level Analysis}.}
    \item The parameters of the logistic normal distribution depend on document-level metadata covariates $\mathbf{x}_d$.
\end{itemize}

For parameter estimation, the STM employs a variational expectation maximization (EM) algorithm, where in the E-step the variational posteriors are updated using a Laplace approximation \citep{wang2013variational, roberts2016model}. In the M-step, the approximated Kullback-Leibler (KL) divergence is minimized with respect to the model parameters.

\section{Modeling Topic-Metadata Relationships in the STM}
\label{Topic-Metadata Relationships in the STM}

The STM produces an approximate posterior distribution of topic proportions. A point estimate can be obtained for example as the mode of this distribution. Topic proportions are often used in subsequent analysis, e.g., for determining their relationship with metadata. We argue that the usual practice of simply regressing point estimates of topic proportions on document-level covariates is not adequate for estimating topic-metadata relationships. This approach ignores that topic proportions are themselves estimates, neglecting much of the information contained in their posterior distribution. In this section, we propose a method to adequately explore the relationship between topic proportions and metadata covariates.

One way to account for the uncertainty in topic proportions is the "method of composition" \cite[p.52;][]{tanner2012tools}, which is a simple Monte Carlo sampling technique. Let $y$ be a random variable with unknown distribution $p(y)$ from which we would like to sample and let $z$ be another random variable with known distribution $p(z)$. If $p(y \vert z)$ is known, we can sample from

\begin{align}
    p(y) = \int p(y \vert z) p(z) dz,
\end{align}

\noindent using the following procedure:

\begin{enumerate}
    \item Draw $z^* \sim p(z)$.
    \item Draw $y^* \sim p(y \vert z^*)$.
\end{enumerate}

\noindent Discarding $z^*$, the resulting $y^*$ are samples from $p(y)$.\footnote{Note that this method is an exact sampling method.}

In \citet{roberts2016model}, the authors employ a variant of the method of composition established by \citet{treier2008democracy}, which uses linear regression to obtain the conditional distribution $p(y \vert z)$. To demonstrate this variant, let $\boldsymbol{\theta}_{\bigcdot k}=(\theta_{1,k}, \dots, \theta_{D,k})^T \in (0,1)^{D}$ denote the proportions of topic $k$ and let $\mathbf{X}:=[\mathbf{x}_1 \vert \dots \vert \mathbf{x}_D]^T$ be the covariates for all $D$ documents. Let further $q(\boldsymbol{\theta}_{\bigcdot k})$ be the approximate posterior distribution of topic proportions given observed documents and metadata, as produced by the STM. The idea now is to repeatedly draw samples $\boldsymbol{\theta}_{\bigcdot k}^*$ from $q(\boldsymbol{\theta}_{\bigcdot k})$ and subsequently perform a regression of each sample $\boldsymbol{\theta}_{\bigcdot k}^*$ on covariates $\mathbf{X}$ to obtain coefficient estimates $\hat{\boldsymbol{\xi}}$. \citet{treier2008democracy} consider the asymptotic distribution of $\hat{\boldsymbol{\xi}}$ as posterior density for $\boldsymbol{\xi}$, i.e., as $p(\boldsymbol{\xi} \vert \boldsymbol{\theta}_{\bigcdot k}^*, \mathbf{X})$. 

That is, the method of composition draws samples from the asymptotic distribution of the 
Maximum Likelihood Estimate (MLE) for the regression parameters. This use of the asymptotic distribution of the MLE can be motivated by the idea that the prior distribution is dominated by the likelihood for larger samples. Therefore, the posterior can be shown to be approximately normal with mean vector equal to the MLE and variance equal to the inverse observed information matrix \citep[see, e.g.,][]{walker1969asymptotic}.

Using samples $\boldsymbol{\xi}^*$ from this distribution $p(\boldsymbol{\xi} \vert \boldsymbol{\theta}_{\bigcdot k}^*, \mathbf{X})$, we can ``predict'' topic proportions $\theta_{pred, k}^{*} = g(\mathbf{x}_{pred}^T \boldsymbol{\xi}^*)$ at new covariate values $\mathbf{x}_{pred}$ ($g$ is the regression response function, e.g., identity function for linear regression). Algorithm \ref{algo_freq} summarizes the method. Note that sampling from the posterior of topic proportions in the first step of Algorithm \ref{algo_freq} accounts for the uncertainty in $\boldsymbol{\theta}_{\bigcdot k}$, while the uncertainty of the regression estimation itself is addressed by sampling from the (asymptotic) distribution of the regression coefficient estimator.

\begin{algorithm*}[ht]
    \SetKwInOut{Input}{Input}
    \SetKwInOut{Output}{Output}
    \textbf{repeat procedure} $m$ \textbf{times}:\\
    \hspace{8pt} Draw $\boldsymbol{\theta}^*_{\bigcdot k} \sim q(\boldsymbol{\theta}_{\bigcdot k})$, where $q$ is the approximate posterior of $\boldsymbol{\theta}_{\bigcdot k}$.\\
    \hspace{8pt} Regress $\boldsymbol{\theta}^*_{\bigcdot k}$ on $\mathbf{X}$; store estimated regression coefficients $\hat{\boldsymbol{\xi}}$ and corresponding covariance matrix.\\
    \hspace{8pt} Draw $\boldsymbol{\xi}^*$ from the (asymptotic) distribution of $\hat{\boldsymbol{\xi}}$.\\
    \hspace{8pt} Predict topic proportions $\theta_{pred, k}^{*} = g(\mathbf{x}_{pred}^T \boldsymbol{\xi}^*)$ at new covariate values $\mathbf{x}_{pred}$.\\
    \textbf{end procedure}
    \caption{Method of composition with frequentist regression}
    \label{algo_freq}
\end{algorithm*}

To visualize topic-metadata relationships, \citet{roberts2016model} generate multiple ``predictions'' $\theta_{pred, k}^{*}$ and calculate empirical quantities such as the mean and quantiles. Calculating mean and credible intervals in such a Bayesian fashion implicitly assumes a (posterior predictive) distribution for $\theta_{pred, k}^{*}$. This distribution, however, directly depends on the regression - which is frequentist as implemented in the \texttt{stm} package. We address this point in detail in Section \ref{bayesian_beta}.  

\section{Methodological Improvements}
\label{Methodological Improvements}

\begin{figure}[ht]
  \centering
    \includegraphics[width=\linewidth]{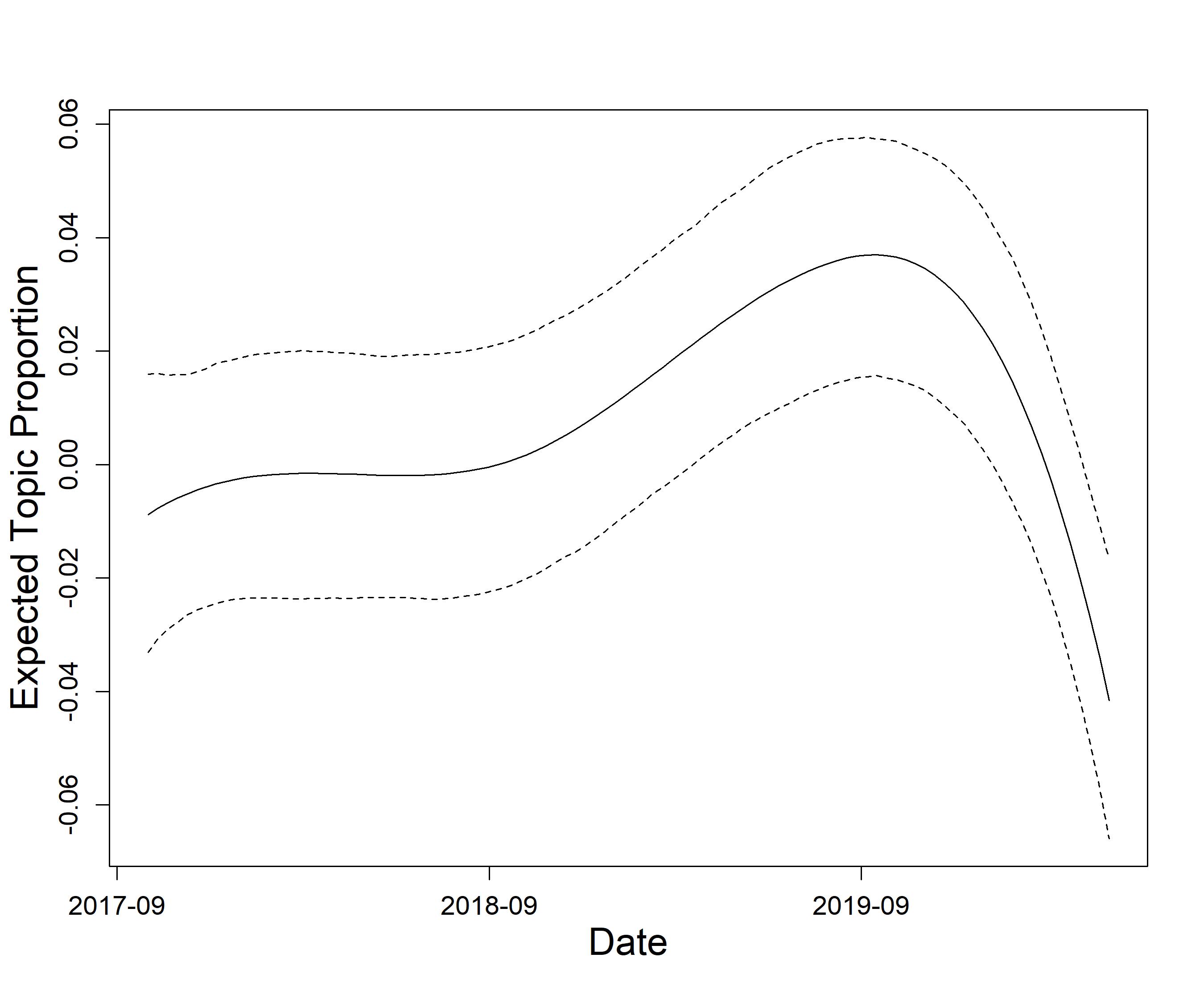}
  \caption{Mean prediction and 95\% confidence intervals for the topic proportion of topic ``Climate Protection'' over time, generated
using \texttt{estimateEffect} from the \textsf{R} package \texttt{stm}.}
  \label{fig:estimateEffect}
\end{figure}

While we agree with performing Monte Carlo sampling of topic proportions in order to integrate over latent variables, we aim to address two inconsistencies:

\begin{itemize}
\item[1.] \textbf{Inadequate modeling of proportions}: The method of composition is implemented in the \textsf{R} package \texttt{stm} via the \texttt{estimateEffect} function, which employs a linear regression in the second step of Algorithm \ref{algo_freq} (implying $g = id$ in the last step). This implementation ignores that topic proportions are naturally restricted to the interval $(0,1)$. As a consequence, when using the \texttt{estimateEffect} function, we frequently observed predicted topic proportions outside of $(0,1)$, as is exemplarily shown for one specific topic-covariate combination in Figure \ref{fig:estimateEffect}.

\item[2.] \textbf{Mixing Bayesian and frequentist methods}: The method of composition used by \citet{treier2008democracy} and \citet{roberts2016model} mixes Bayesian and frequentist methods. As described in Section \ref{Topic-Metadata Relationships in the STM}, a frequentist regression is used inside the method of composition, yet estimates are obtained in a Bayesian manner via calculation of empirical mean and quantiles. Recall that according to \citet{treier2008democracy}, $\boldsymbol{\xi}^*$ can be considered a sample from the posterior of regression coefficients. However, the coefficients resulting from a frequentist regression do not have any distribution because the frequentist framework assumes them to be fixed parameters. As a consequence, one cannot sample from the distribution of regression coefficients, which is why \citet{treier2008democracy} sample $\boldsymbol{\xi}^*$ from the distribution of coefficient \textit{estimators}. This distribution, however, only exists by making frequentist assumptions.
\end{itemize}
In Sections \ref{frequentist_beta} and \ref{bayesian_beta} below we further discuss these problems and present corrections and alternatives, all of which are implemented in the \textsf{R} package \texttt{stmprevalence}.
\footnote{Available at \href{https://github.com/PMSchulze/stmprevalence}{https://github.com/PMSchulze/stmprevalence}.}

\subsection{Frequentist Beta Regression}
\label{frequentist_beta}

As noted above, the linear regression approach is often used carelessly in the literature, neglecting that topic proportions are non-negative by definition. \citet{farrell2016corporate} and \citet{moschella2019central}, for instance, produce figures containing negative expected topic proportions, while \citet{cho2017anchoring}, \citet{chandelier2018content}, \citet{bohr2018key}, and \citet{heberling2019changing} display confidence bands partly covering negative values.

Therefore, we correct the approach employed within the \texttt{stm} package by replacing the linear regression with a regression model that assumes a dependent variable in the interval $(0,1)$. As shown by \citet{atchison1980logistic}, the Dirichlet distribution is well suited to approximate a logistic normal distribution, though inducing less interdependence among the different topics. When employing a Dirichlet distribution, the univariate marginal distributions are Beta distributions. We thus perform a separate Beta regression for each topic proportion on $\mathbf{X}$, using a logit-link.\footnote{Note that the distribution of regression coefficient estimators is asymptotically normal for Beta regression \cite[p.17;][]{ferrari2004Beta}.} This approach now again corresponds to Algorithm \ref{algo_freq}, but with $g$ being the logistic sigmoid function in this case.\footnote{While runtime for estimating Beta regressions is considerably longer in relative terms, it is still short in absolute terms, which is why runtime concerns can be disregarded for the practical use of our approach.}

\subsection{Bayesian Beta Regression}
\label{bayesian_beta}

\citet{treier2008democracy} and the authors of the STM consider $\boldsymbol{\xi}^*$ to be samples from the posterior of regression coefficients. While it is possible to view frequentist regression from a Bayesian perspective, it implies assuming a uniform prior distribution for regression coefficients $\boldsymbol{\xi}$ - which is rather implausible. More generally, the mixing of Bayesian and frequentist frameworks within the method of composition lacks a theoretical foundation, especially when employing an \textit{asymptotic} distribution of regression coefficient estimators. This applies to the model of \citet{treier2008democracy} as well as to the Beta regression presented in Section \ref{frequentist_beta}. Furthermore, note that when using a frequentist regression, the estimated uncertainty is with respect to the prediction of the mean of topic proportions. However, when exploring topic-metadata relationships it might be preferable to examine the variation of individual topic proportions among documents at different values of metadata covariates.

\begin{algorithm*}[ht]
    \SetKwInOut{Input}{Input}
    \SetKwInOut{Output}{Output}
    \textbf{repeat procedure} $m$ \textbf{times}:\\
    \hspace{8pt} Draw $\boldsymbol{\theta}^*_{\bigcdot k} \sim q(\boldsymbol{\theta}_{\bigcdot k})$, where $q$ is the approximate posterior of $\boldsymbol{\theta}_{\bigcdot k}$.\\
    \hspace{8pt} Perform a Bayesian Beta regression of $\boldsymbol{\theta}^*_{\bigcdot k}$ on $\mathbf{X}$ using normal priors centered around zero.\\
    \hspace{8pt} Draw $\theta_{pred, k}^{*} \sim p(\boldsymbol{\theta}_{pred, k} \vert \boldsymbol{\theta}^*_{\bigcdot k}, \mathbf{X}, \mathbf{x}_{pred})$, i.e., conditional on sample $\boldsymbol{\theta}^*_{\bigcdot k}$.\\
    \textbf{end procedure}
    \caption{Method of composition with Bayesian Beta regression}
    \label{algo_bayes}
\end{algorithm*}

\noindent Therefore, we propose to replace the frequentist regression in Algorithm \ref{algo_freq} by a Bayesian Beta regression with normal priors centered around zero. This enables modeling topic-metadata relationships in a fully Bayesian manner while preserving the methodological improvements from Section \ref{frequentist_beta}. Algorithm \ref{algo_bayes} summarizes this approach. By drawing $\theta_{pred, k}^{*}$ at covariate values $\mathbf{x}_{pred}$, we obtain samples from the posterior predictive distribution
\begin{align}
    p(\boldsymbol{\theta}_{pred, k}& \vert \boldsymbol{\theta}^*_{\bigcdot k}, \mathbf{X}, \mathbf{x}_{pred}) = \\& \int p(\boldsymbol{\theta}_{pred, k} \vert \mathbf{x}_{pred}, \boldsymbol{\xi}) p(\boldsymbol{\xi}  \vert \boldsymbol{\theta}_{\bigcdot k}^*, \mathbf{X}) d\boldsymbol{\xi},
\end{align}
where $p(\boldsymbol{\xi}  \vert \boldsymbol{\theta}_{\bigcdot k}^*, \mathbf{X})$ denotes the posterior distribution of regression coefficients. This allows displaying the (predicted) variation of topic proportions at different covariate levels. As before, quantities of interest, such as the mean and quantiles, are obtained by averaging across samples; now, however, these samples are generated within a fully Bayesian framework.

\section{Application\protect\footnote{Source code available at \href{https://github.com/PMSchulze/topic-metadata-stm}{https://github.com/PMSchulze/\\topic-metadata-stm}.}}
\label{Application}

In this section, we first apply the STM to German parliamentarians' Twitter data and subsequently demonstrate both the original (\texttt{stm}) and our new method (\texttt{stmprevalence}) to explore topic-metadata relationships. Here, we chose to apply the STM in particular for illustrative purposes, because of its flexibility and its relevance in the social sciences. We would like to emphasize again, however, that our methods work with any other topic model, such as LDA or CTM, as long as it produces an (approximate) posterior distribution of topic proportions. This is because our methods focus on the step subsequent to the estimation of a topic model, i.e., on the exploration of relationships between previously estimated topic proportions and metadata covariates.

\subsection{Data\protect\footnote{Raw data: \href{https://figshare.com/s/7a728fcb6d67a67fc3d6}{https://figshare.com/s/7a728fcb6d67a67fc3d6}.}}
\label{Data}

For all German MPs during the 19th election period (starting on September 24, 2017), we gathered personal information such as name, party affiliation, and electoral district from the official parliament website as well as Twitter profiles from the official party websites, using \textit{BeautifulSoup} \citep{richardson2007beautiful}. Next, after excluding MPs without a public Twitter profile, we used \textit{tweepy} \citep{roesslein2020tweepy} to scrape all tweets by German MPs from September 24, 2017 through April 24, 2020. We also gathered socioeconomic data, such as GDP per capita and unemployment rate, as well as 2017 election results on an electoral-district level. Text preprocessing, such as transcription of German umlauts, removal of stopwords, and word-stemming, was performed with \textit{quanteda} \citep{quanteda}.\footnote{An in-depth discussion of topic model preprocessing and its application to Twitter data can be found in \citet{lucas2015computer}.}

We define a document as the concatenation of an individual MP's tweets during a single calendar month in order to achieve sufficient document length. Our final data set includes 10,998 monthly MP-level documents, each one associated with 90 covariates.

\subsection{Model Fitting and Global-level Analysis}
\label{Model Fitting and Global-level Analysis} 

Before fitting the STM, we need to decide on the number of topics, $K$. To do so, we use the following four model evaluation metrics: \textit{held-out likelihood}, \textit{semantic coherence}, \textit{exclusivity}, and \textit{residuals}. The held-out likelihood approach is based on document completion. The higher the held-out likelihood, the more predictive power the model has on average \citep{wallach2009evaluation}. Semantic coherence means that words characterizing a specific topic also appear together in the same documents \citep{mimno2011optimizing}. Exclusivity, on the other hand, indicates to which degree words characterizing a given topic \textit{only} occur in that topic. Finally, the residuals metric, which is based on residual dispersion, indicates a (potentially) insufficiently small value of $K$ whenever the residual dispersion is larger than one \citep{taddy2012estimation}.

\begin{figure*}[ht]
  \centering
  \begin{subfigure}[b]{0.49\linewidth}
    \includegraphics[width=\linewidth]{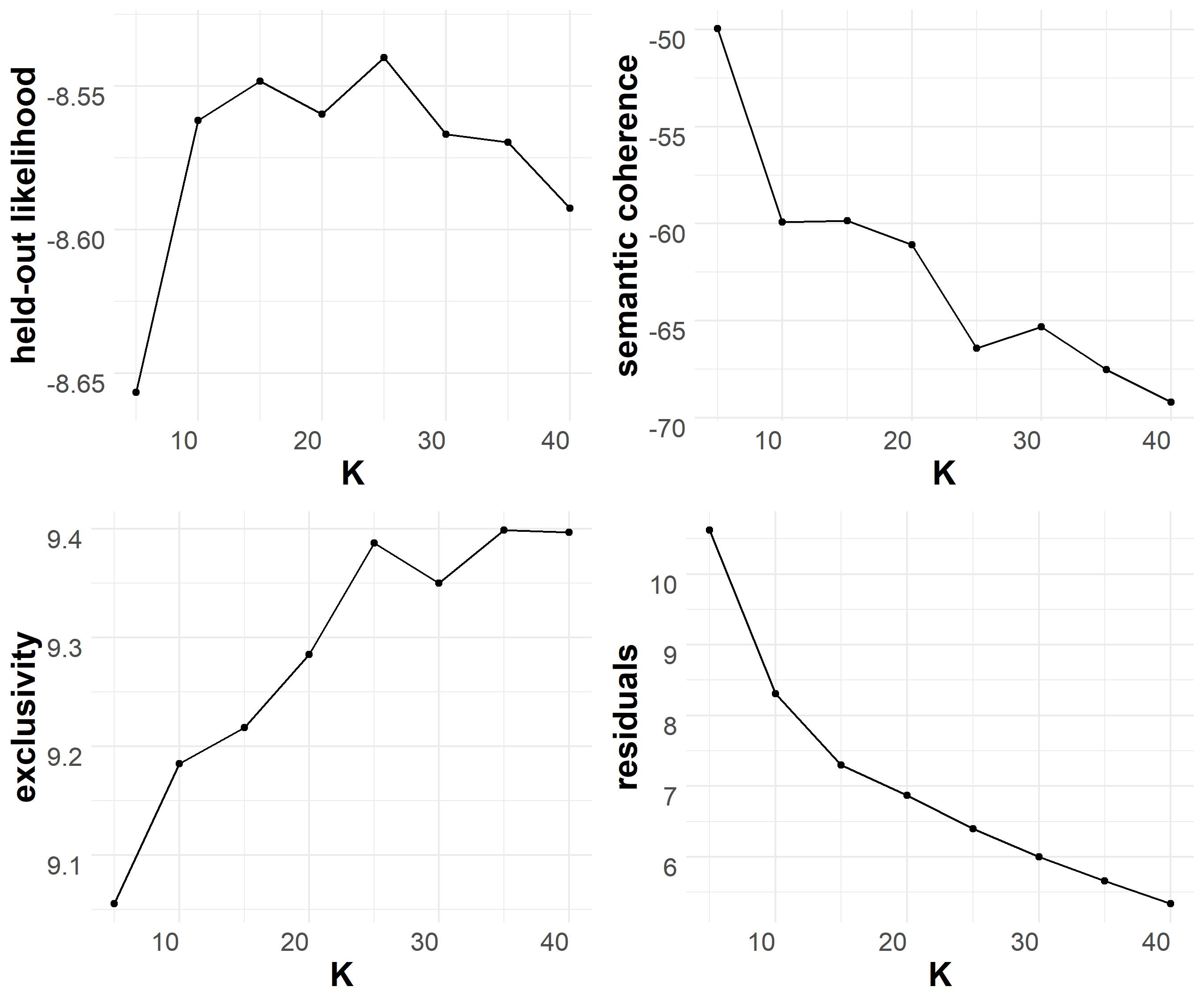}
  \end{subfigure}
  \begin{subfigure}[b]{0.49\linewidth}
    \includegraphics[width=\linewidth]{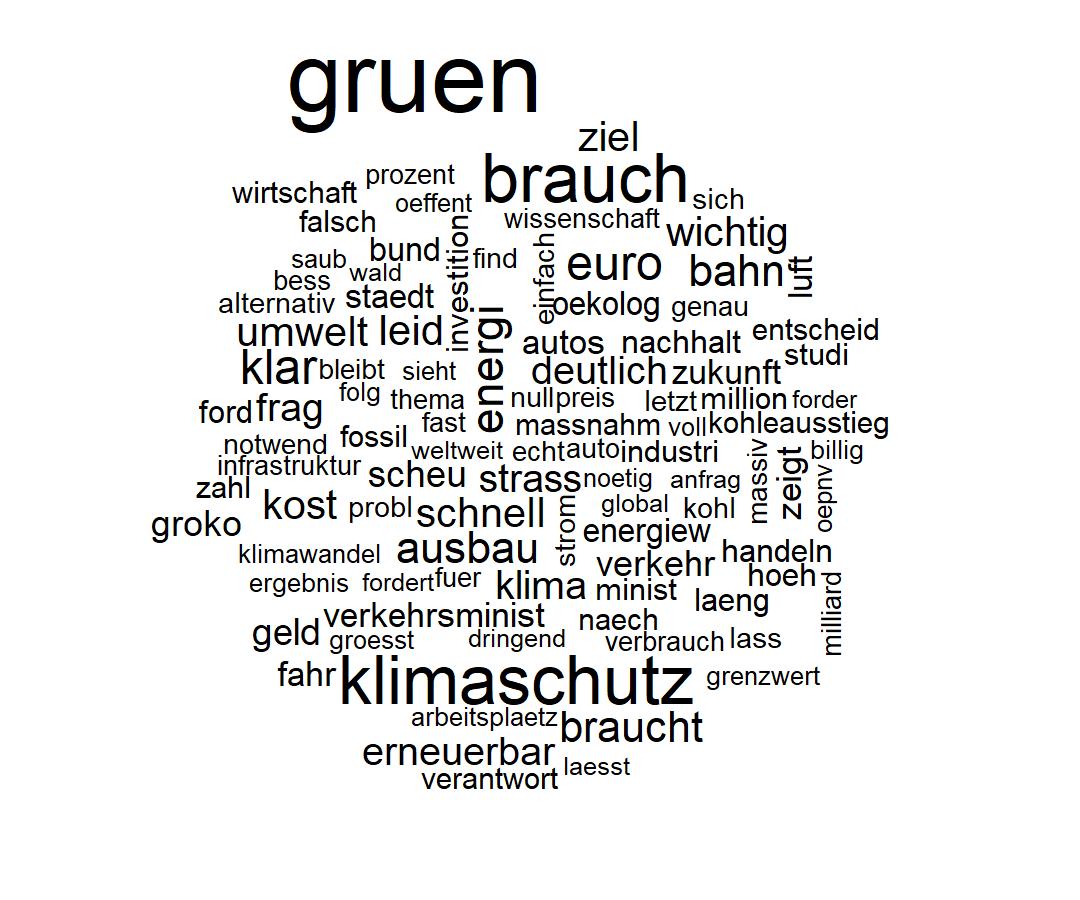}
  \end{subfigure}
  \caption{Left: Model evaluation metrics for hyperparameter $K$ (number of topics). Right: Word cloud for the topic labeled as ``Climate Protection''.}
  \label{fig:searchK_and_wordcloud}
\end{figure*}

The left part of Figure \ref{fig:searchK_and_wordcloud} shows these four metrics for a grid of $K$ between five and 40 with step size five. Both $K=15$ and $K=20$ seem to be good choices. Given the better interpretability for models with fewer topics, we choose $K = 15$.

After fitting the model, we label all topics manually with human interpretable labels; to do so, we use word clouds and top words (see Figure \ref{fig:searchK_and_wordcloud} (right panel) and Appendix \ref{a:further_cloude}). Throughout this work, we consider the topics ``Climate Protection'', ``Right/Nationalist'', ``Social/Housing'', and ``Europe'' for illustration, in particular the first one. To obtain an overview of the model output, different global-level analyses are conducted, such as inspecting global topic proportions $\bar{{\theta}}_k = \frac{1}{D}\sum_{d=1}^{D}\theta_{d,k}$ or creating a network graph.

\subsection{Topic-Metadata Relationships}
\label{Topic-Metadata Relationships}

\begin{figure*}[ht]
  \centering
  \begin{subfigure}[b]{0.49\linewidth}
    \includegraphics[width=\linewidth]{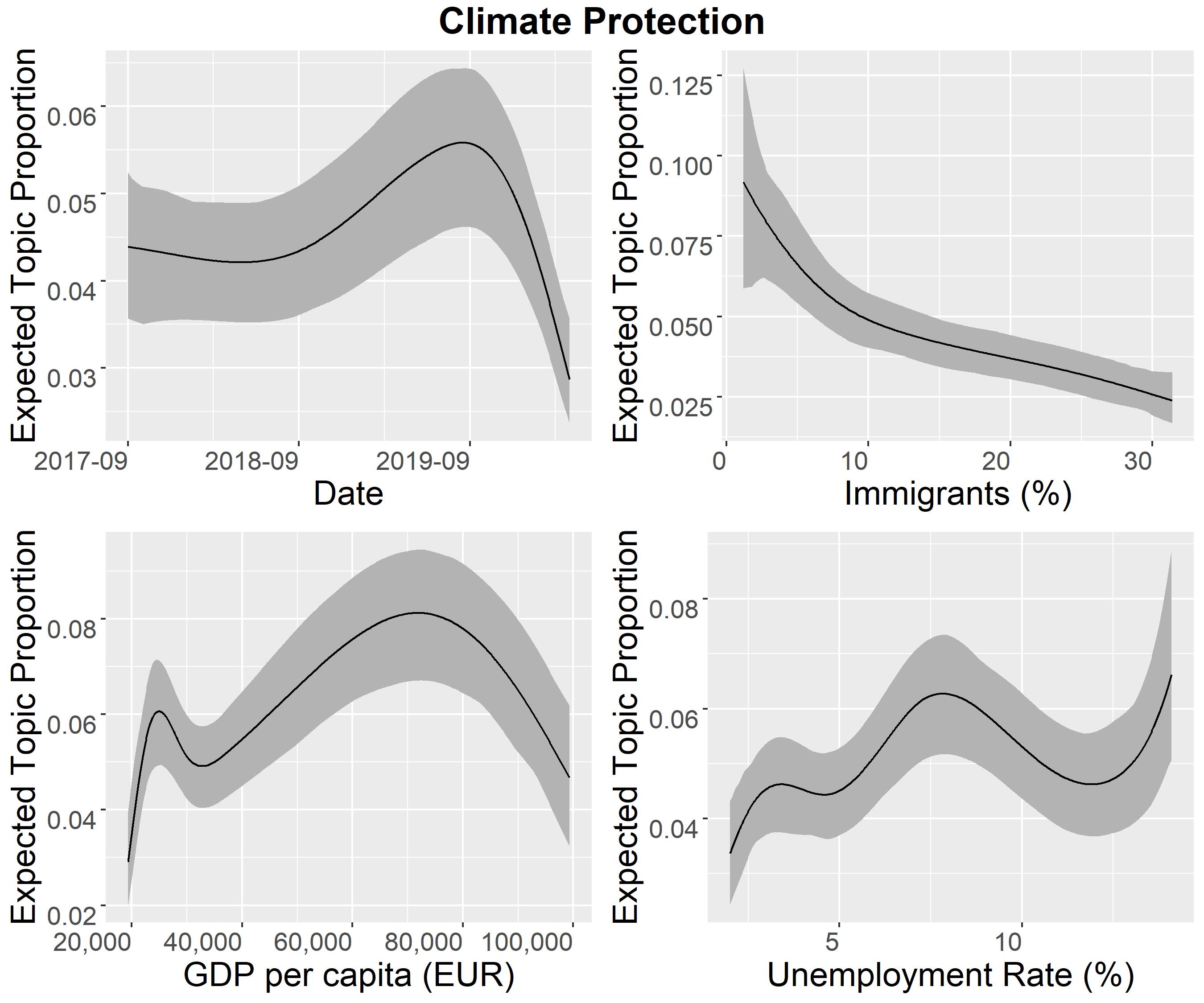}
  \end{subfigure}
  \begin{subfigure}[b]{0.49\linewidth}
    \includegraphics[width=\linewidth]{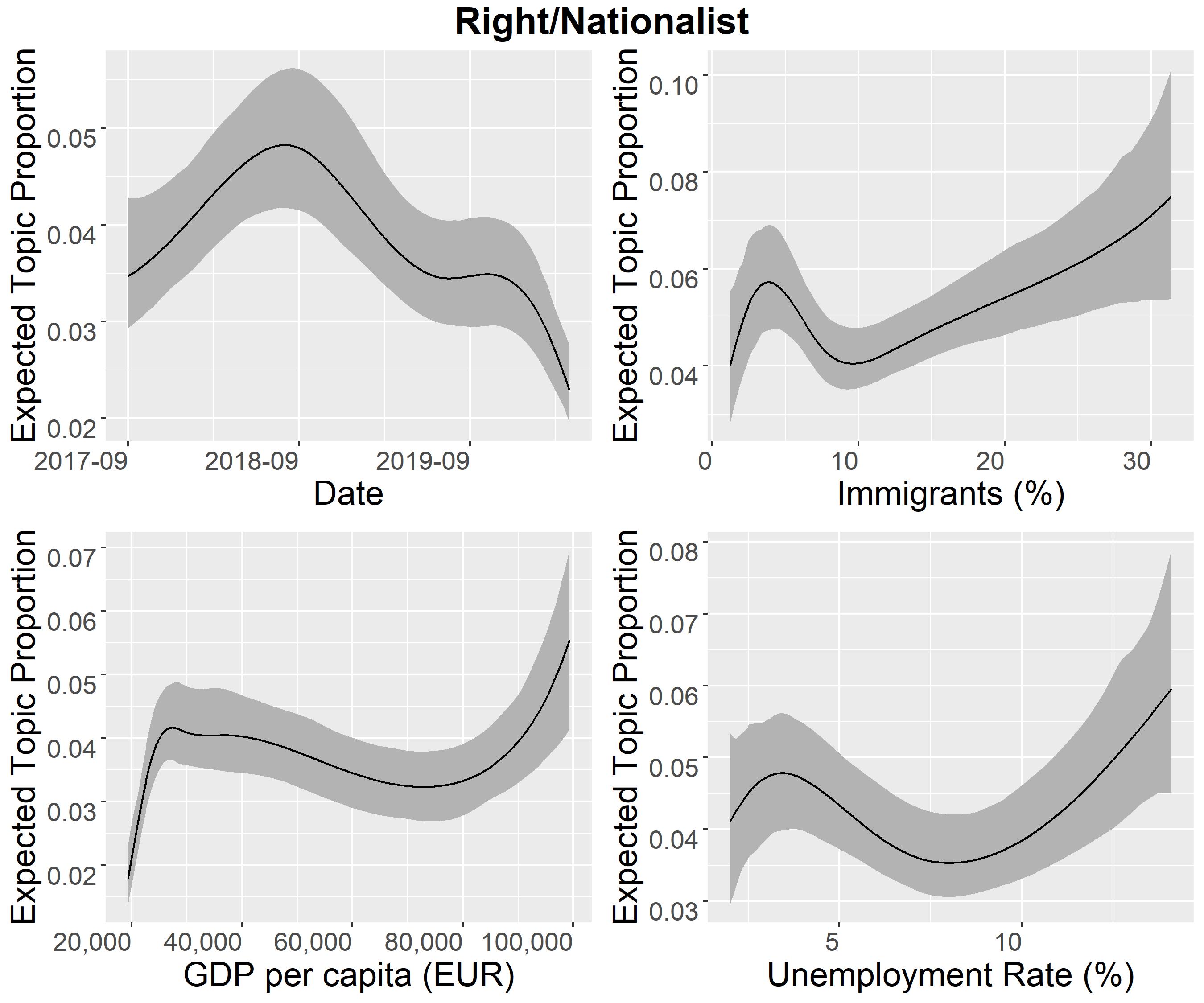}
  \end{subfigure}
  \begin{subfigure}[b]{0.49\linewidth}
    \includegraphics[width=\linewidth]{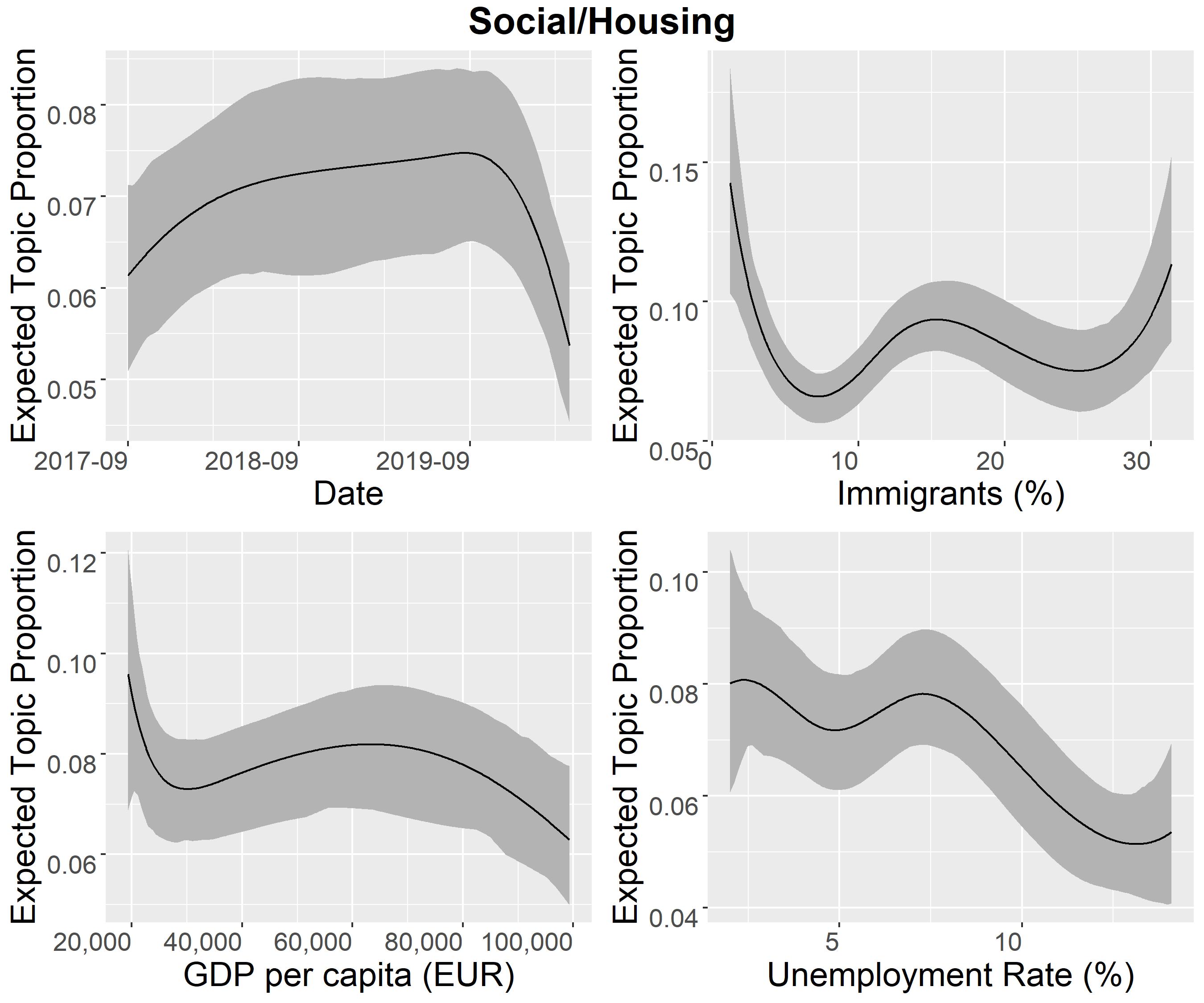}
  \end{subfigure}
  \begin{subfigure}[b]{0.49\linewidth}
    \includegraphics[width=\linewidth]{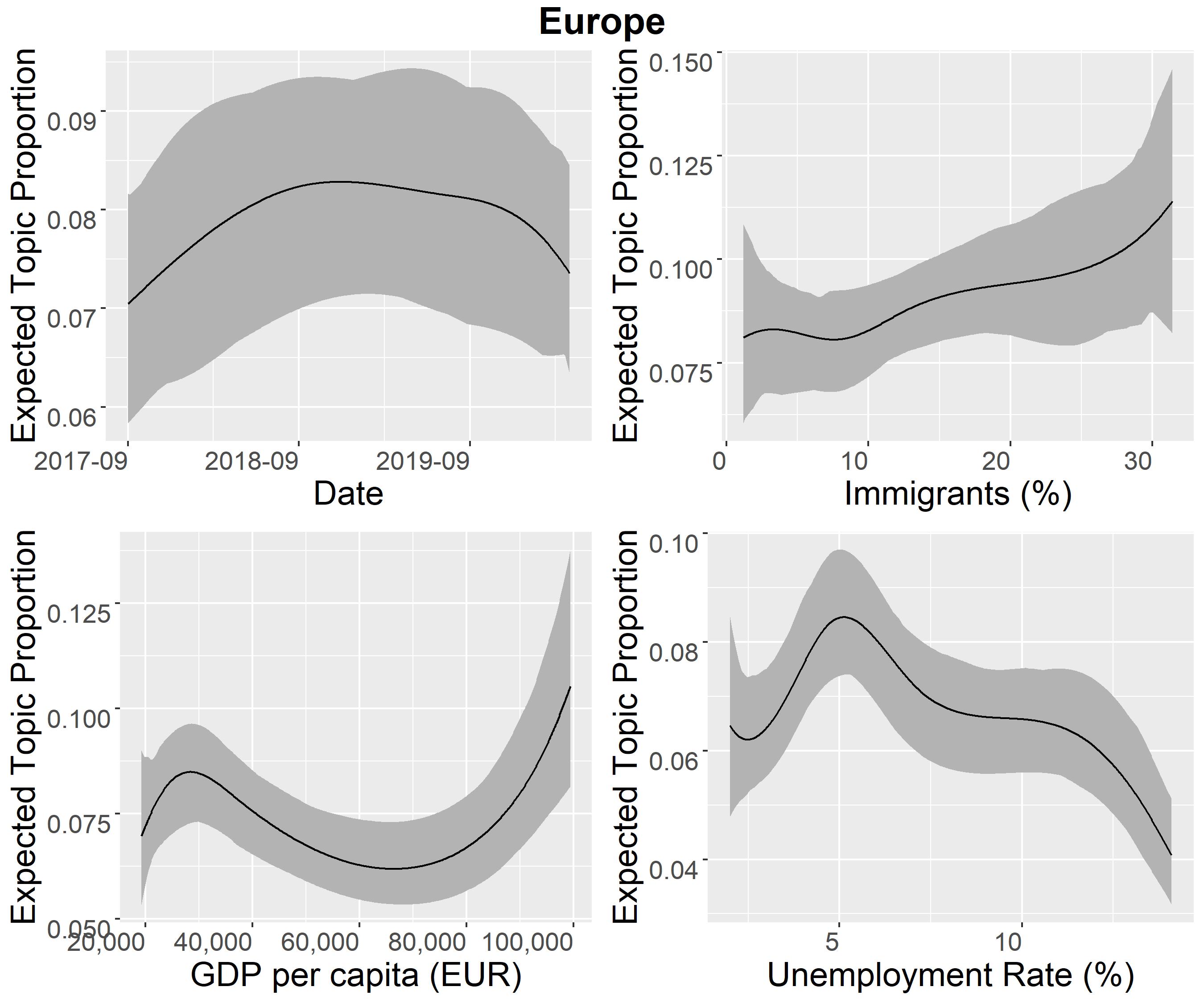}
  \end{subfigure}
  \caption{Mean prediction and 95\% confidence intervals for the topic proportion of topics ``Climate Protection'', "Right/Nationalist", "Social/Housing", and "Europe" for different document-level covariates, obtained using a frequentist Beta regression from the \textsf{R} package \texttt{stmprevalence}.}
  \label{fig:frequentist}
\end{figure*}

\begin{figure*}[ht]
  \centering
  \begin{subfigure}[b]{0.49\linewidth}
    \includegraphics[width=\linewidth]{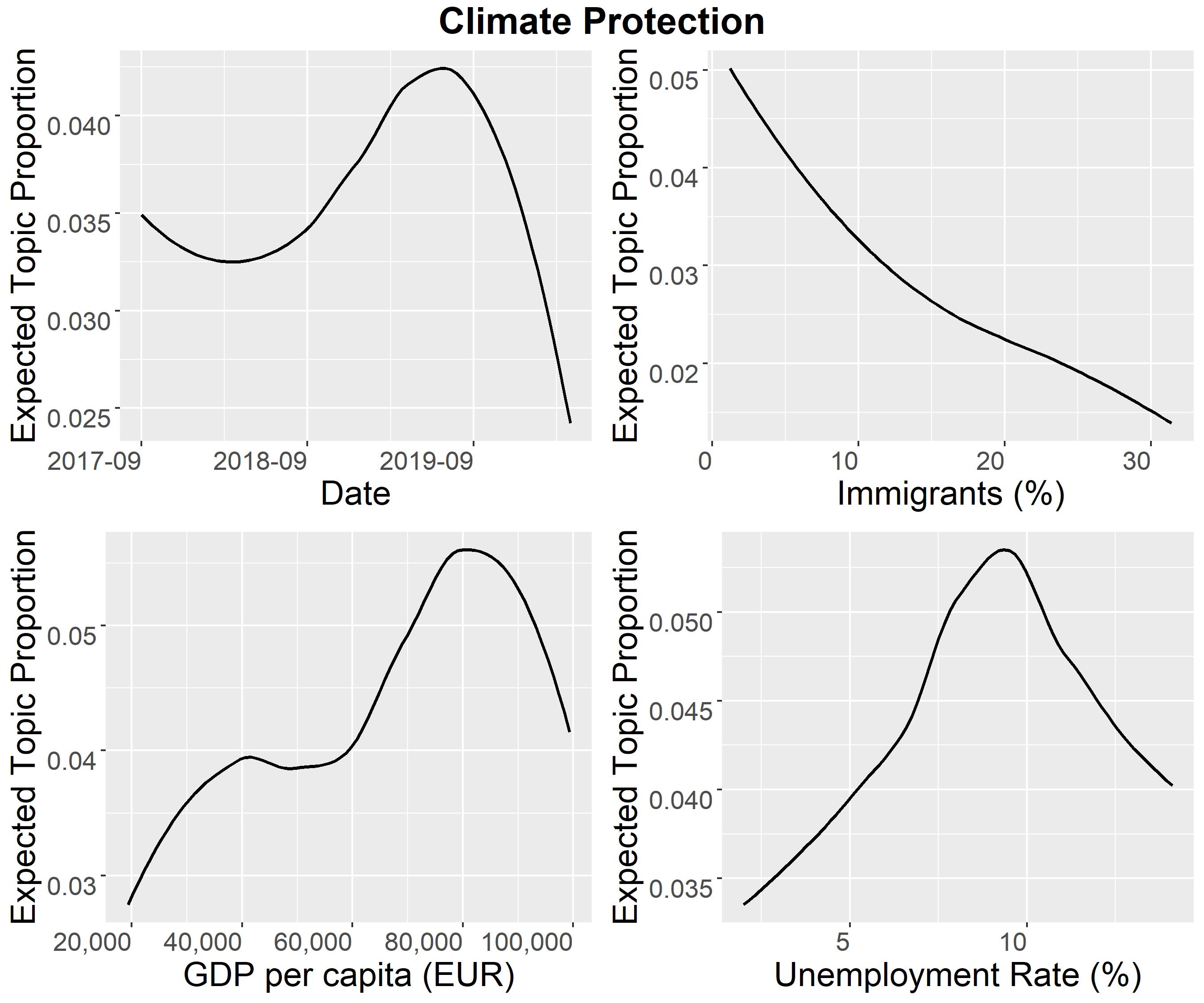}
  \end{subfigure}
  \begin{subfigure}[b]{0.49\linewidth}
    \includegraphics[width=\linewidth]{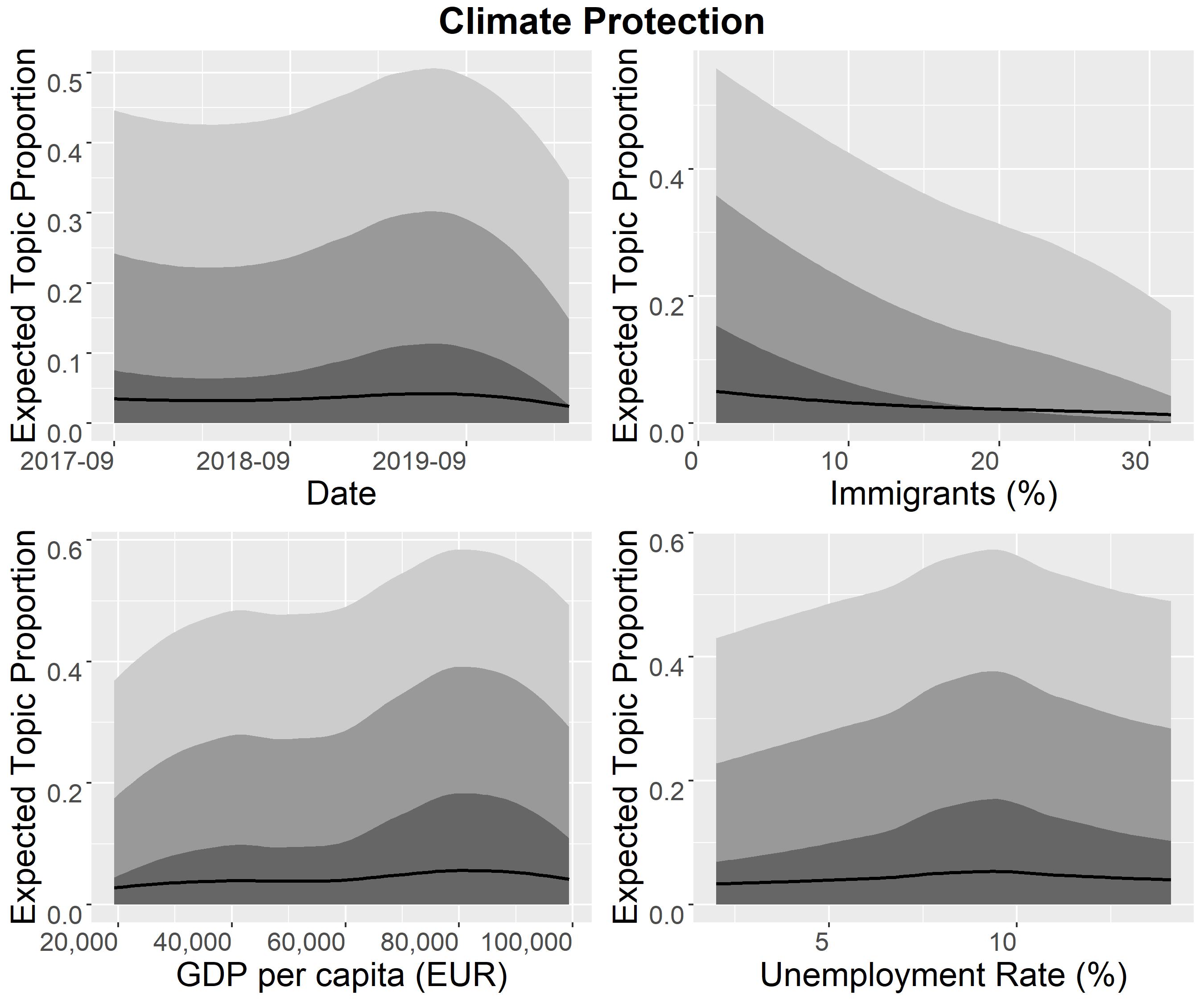}
  \end{subfigure}
  \caption{Left: Mean prediction for the topic proportion of topic ``Climate Protection'' for different document-level covariates, obtained using a Bayesian Beta regression from the \textsf{R} package \texttt{stmprevalence}. Right: 95\% (light grey), 90\% (grey), and 85\% (dark grey) quantiles of the posterior predictive distribution for the topic proportion of topic ``Climate Protection''.}
  \label{fig:Bayesian}
\end{figure*}

Moving from global- to document-level, we now visualize relationships between document-level topic proportions $\theta_{d,k}$ and covariates $\mathbf{x}_d$. In particular, we examine the extent to which German MPs discussed the abovementioned topics over time and in relation to several socioeconomic variables regarding their respective electoral districts. These relationships were estimated by regressing the previously estimated topic proportions on metadata covariates, using either the linear regression-based method of composition (see Fig. \ref{fig:estimateEffect}) or our Beta regression-based methods (see Fig. \ref{fig:frequentist} and \ref{fig:Bayesian}).\footnote{Again, note that the topic proportions could alternatively have been estimated via, e.g., LDA or CTM. Our methods concern the subsequent step, i.e., estimating topic-metadata relationships, and are unrelated to the topic model choice.} 

For all regressions, we choose the same linear predictor, containing the date of the Twitter posts, the MP-level categorical covariates \textit{political party affiliation} and \textit{federal state}, as well as the electoral district-level continuous socioeconomic covariates \textit{immigration share}, \textit{GDP per capita}, and \textit{unemployment rate}; the effects of the latter three, due to being continuous, are estimated as smooth functions using B-splines.



To demonstrate the shortcomings of the approach implemented in the \texttt{stm} package, we first apply the \texttt{estimateEffect} function to produce ``na\"{i}ve'' estimates for the relationship between estimated topic proportions and document-level covariates. Figure \ref{fig:estimateEffect} shows the estimated proportion of the topic ``Climate Protection'' over time, peaking during the UN Climate Action Summit 2019 held in September 2019. Importantly, notice that \texttt{estimateEffect} produces predicted topic proportions outside of $(0,1)$. This is due to using a linear regression, which places no restrictions on the range of the dependent variable. 

Next, we evaluate the results when replacing the linear regression by a Beta regression, which restricts the dependent variable to the $(0,1)$-interval.

Figure \ref{fig:frequentist} consists of four panels, one for each topic, each panel being made up of four (sub)plots. The top left plot in the top left panel corresponds to the time trend of the climate protection topic. It shows that the overall trend over time is similar to the one in Figure \ref{fig:estimateEffect}, yet the range is shifted upwards and no negative values are estimated. The three remaining plots of the top left panel depict the relationship of the climate protection topic with the socioeconomic covariates immigration, GDP per capita, and unemployment as measured at the electoral district-level. First, note that only non-negative values are obtained - as desired. Regarding GDP per capita, we notice an increase in the relevance of the climate protection topic until around EUR 70k, yet for very high income electoral districts this trend is reversed. The unemployment rate shows an ambiguous relationship, with rather large fluctuations. Finally, the higher the share of immigrants in an electoral district, the less frequently the district's MPs tend to discuss climate-related subjects on average.

However, one might suspect that this negative relationship between climate protection relevance and immigration is the consequence of spurious correlation: one immigration-related topic might simply be suppressing all other topics.\footnote{Recall that topic proportions must sum to $1$, so an increase in the proportion of one topic mechanically decreases the relevance of all other topics.} To investigate this, and also in order to evaluate our approach more broadly, we consider three further topics, ``Right/Nationalist'', ``Social/Housing'', and ``Europe''. Actually, the frequency of the ``Right/Nationalist'' topic increases as electoral district-level immigrant share increases, yet a similar association can also be found for the Europe-related topic; for the topic regarding social issues and housing, no clear trend is recognizable. This leads us to conclude that the negative association between the relevance of the climate protection topic and the immigration share is not only an effect of the mechanics of compositional data such as topic proportions.

Regarding time, the social and European topics do not show any temporal trend, whereas the nationalist topic clearly peaks around September 2018. As for GDP per capita and unemployment rate, only few more or less clear trends can be recognized, such as the decrease in the relevance of the European as well as the social topic with increasing unemployment rate. However, while some interesting and reasonable patterns emerge, we do caution against (quantitative) over-interpretation of the observed patterns.

Finally, we display the results from the fully Bayesian approach discussed in Section \ref{bayesian_beta}, though here we only focus on the climate protection topic for the sake of brevity. As can be seen in the left plot of Figure \ref{fig:Bayesian}, the predicted progressions of mean topic proportions at different covariate values are mostly similar to those obtained with the frequentist Beta regression, yet the range is compressed and shifted downwards. In addition to the empirical mean, the right plot of Figure \ref{fig:Bayesian} depicts different empirical quantiles of the posterior predictive distribution of topic proportions. Here we can see that topic proportions at different covariate values vary starkly for different MPs. More generally, we find that a fully Bayesian approach enables a much more comprehensive analysis of topic-metadata relationships because it allows for displaying the variation of individual topic proportions observed in the data.

\section{Conclusion}

Nowadays, large-scale unstructured text from a wide variety of fields is publicly available on social media and various other forms of online appearances. Topic modeling plays an important role in the extraction of specific information from such data. At the same time, researchers - in particular from the social sciences - increasingly move beyond purely exploratory topic analyses, wishing to associate identified topics with metadata. In order to investigate topic-metadata relationships while accounting for the probabilistic nature of topic proportions, the \textsf{R} package \texttt{stm} implements repeated linear regressions of sampled topic proportions on metadata covariates using the method of composition. 

In this paper, we identify two main inconsistencies of this original implementation: the inadequate modeling of proportions via linear regression, allowing topic proportions to take on values outside of $(0,1)$; and the mixing of frequentist regression with Bayesian computations of empirical quantities. We propose improvements to both shortcomings: the more appropriate Beta regression to account for the distributional nature of topic proportions; and a fully Bayesian approach to replace the current mixture of frequentist and Bayesian methods within the method of composition. 

We illustrate our proposed improvements by first applying the STM to a data set containing Twitter posts by German MPs and subsequently employing our methods to estimate relationships between estimated topic proportions and MP-level metadata covariates. It is important to note that our methods merely concern the second-step estimation of topic-metadata relationships and are thus equally applicable to other topic models and beyond.

\section*{Limitations and Outlook}

There are some limitations to our approach, which in turn give rise to future research. Regarding the application case presented in this paper, the relationship with Twitter-related metadata such as retweets or likes would be interesting - especially because such metadata would be actively influenced by the topics of the tweets, whereas the socioeconomic covariates used here are of a more explanatory nature. Unfortunately, Twitter-related metadata are not contained in the data set. Another use case-related aspect is the document length. Longer documents are beneficial for topic models such as the STM in general, yet in our specific case hamper the content-related interpretability of the resulting ``tweet documents''. We experimented extensively with different document lengths, including days and weeks, but finally came to the conclusion that aggregating tweets at a monthly interval constitutes the best compromise between content-related interpretability and sufficient  text length.

Both frequentist and Bayesian Beta regression are well established approaches in the statistical literature, necessarily implying a lower degree of methodological novelty of our approach. However, the correct modeling and illustration of topic-metadata relationships and the corresponding uncertainty is of paramount importance: because of the enormous popularity of topic models such as the STM and the fact that conclusions drawn from a misspecified model can be (substantially) misleading (cf. Appendix \ref{a:implausible}).

Several possibilities exist to build upon our exploratory methods. For instance, our approach could be used in combination with MCMC-based methods in order to make inference in a Bayesian setting. If the goal is to make causal inference beyond exploratory purposes, one must take into account that the estimation of topic proportions induces additional dependence across documents. Developing methods to identify underlying causal mechanisms is the subject of current research \citep[e.g.,][]{egami2018make}.

\bibliography{anthology,custom}
\bibliographystyle{acl_natbib}

\clearpage

\appendix

\section{Exemplary figures with implausible predictions}\label{a:implausible}

To demonstrate the importance of our proposed corrections of the STM, we collected figures from a selection of research papers where using the original implementation led to implausible estimates. Due to copyright issues, however, we do not show them here but instead merely reference them, along with a short description of how the uncorrected method of composition produces implausible results in the respective cases.

\begin{itemize}
    \item \citet{cho2017anchoring}, p.10, Fig. 10 (actually p.125): negative confidence bands for covariate effects \href{https://ieeexplore.ieee.org/document/8585665}{https://doi.org/10.1109/VAST.2017.8585665                   }
    \item \citet{bohr2018key}, p.9, Fig. 9: negative confidence bands \textit{and} negative covariate effects \href{https://doi.org/10.1080/23251042.2017.1393863                 }{https://doi.org/10.1080/23251042.2017.1393863               }
    \item \citet{moschella2019central}, p.11, Fig. 2 (actually p.523): negative confidence bands \textit{and} negative covariate effects \href{https://doi.org/10.1111/padm.12543                 }{https://doi.org/10.1111/padm.12543}
    \item \citet{chandelier2018content}, p.6, Fig. 2 (actually p.259) : negative confidence bands for covariate effects \href{https://doi.org/10.1016/j.biocon.2018.01.029          }{https://doi.org/10.1016/j                .biocon.2018.01.029}
    \item \citet{heberling2019changing}, p.8, Fig. 5 (actually p.819) : negative covariate effects, \href{https://doi.org/10.1093/biosci/biz094       }{https://doi.org/10.1093/biosci/biz094}
\end{itemize}

\noindent Finally, another example of confidence bands of topic proportions becoming negative when using the \texttt{estimateEffect} function is Figure 7 (p. 20) of the vignette of the \texttt{stm} package. In the README file of our \texttt{stmprevalence} package, we reproduce this figure and furthermore show how the uncertainty estimation is corrected when using our approaches.







\section{Word clouds and top words for selected topics}
\label{a:further_cloude}

The top words for the four topics ``Climate Protection'', ``Right/Nationalist'', ``Social/Housing'', and ``Europe'', which are used for illustration in Figure \ref{fig:frequentist}, are shown in Table \ref{tab:top_words} below.

\begin{table*}[htbp]
  \centering
  \begin{tabular}{|l|l|l|l|l|l|}
    \hline
    \textbf{Topic}          & \textbf{Word 1}  & \textbf{Word 2}  & \textbf{Word 3}  & \textbf{Word 4} & \textbf{Word 5}  \\ \hline
    Climate Protection      & grün            & klimaschutz      & brauch           & klar            & euro             \\ \hline
    Right/Nationalist       & bürg            & link             & merkel           & frau            & sich             \\ \hline
    Social/Housing          & sozial          & miet             & kind             & arbeit          & brauch           \\ \hline
    Europe                  & europäisch      & wichtig          & europa           & international   & thank            \\ \hline
  \end{tabular}
  \caption{Top five words (in terms of absolute frequency across all text documents) within the topics “Climate Protection”, “Right/Nationalist”, “Social/Housing”, and “Europe”.}
  \label{tab:top_words}
\end{table*}

\noindent The word cloud for the ``Climate Protection'' topic has already been shown in Figure \ref{fig:searchK_and_wordcloud} (right panel). Figure \ref{fig:wordclouds} below shows the word clouds for the topics ``Right/Nationalist'', ``Social/Housing'', and ``Europe'', respectively.

\begin{figure*}[ht]
  \centering
  \begin{subfigure}[b]{0.6\linewidth}
    \includegraphics[width=\linewidth]{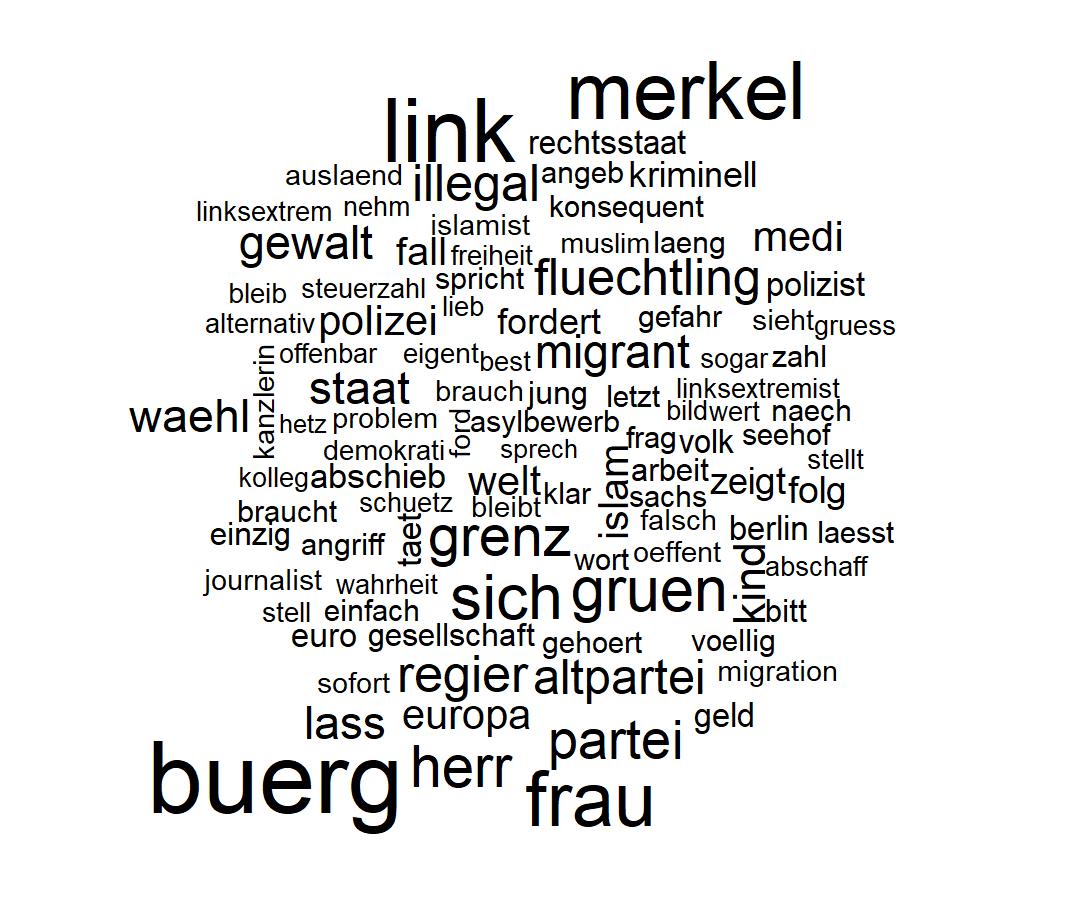}
  \end{subfigure}
  \begin{subfigure}[b]{0.6\linewidth}
    \includegraphics[width=\linewidth]{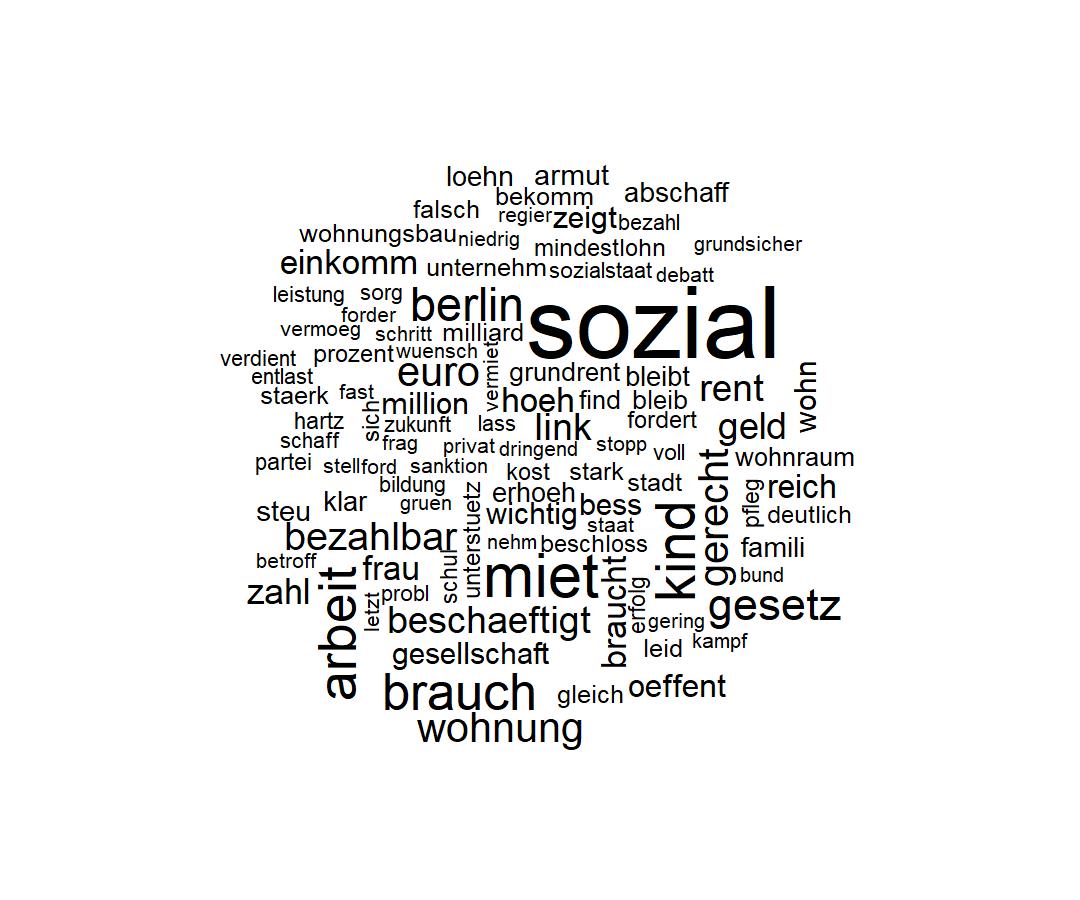}
  \end{subfigure}
  \begin{subfigure}[b]{0.6\linewidth}
    \includegraphics[width=\linewidth]{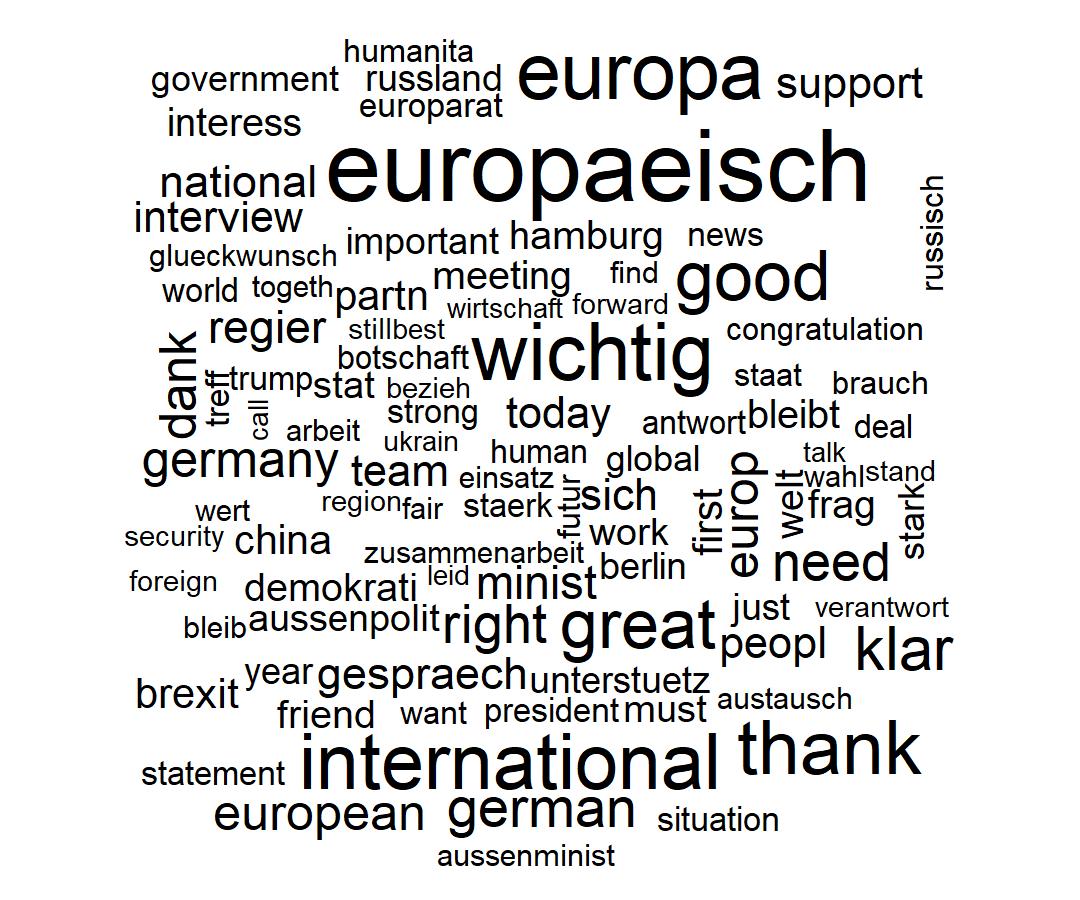}
  \end{subfigure}
  \caption{Word clouds for the topics ``Right/Nationalist'' (top), ``Social/Housing'' (center), and ``Europe'' (bottom).}
  \label{fig:wordclouds}
\end{figure*}

\end{document}